# A Causal Bayesian Model for the Diagnosis of Appendicitis

*Stanley M. Schwartz, Jonathan Baron, and John R. Clarke.*

Department of Psychology
University of Pennsylvania
Philadelphia, PA 19104

GRASP Laboratory
University of Pennsylvania
Philadelphia, PA 19104

Department of Surgery
Medical College of Pennsylvania
Philadelphia, PA 19129

## ABSTRACT

The causal Bayesian approach is based on the assumption that effects (e.g., symptoms) that are not conditionally independent with respect to some causal agent (e.g., a disease) are conditionally independent with respect to some intermediate state caused by the agent, (e.g., a pathological condition). This paper describes the development of a causal Bayesian model for the diagnosis of appendicitis. The paper begins with a description of the standard Bayesian approach to reasoning about uncertainty and the major critiques it faces. The paper then lays the theoretical groundwork for the causal extension of the Bayesian approach, and details specific improvements we have developed. The paper then goes on to describe our knowledge engineering and implementation and the results of a test of the system. The paper concludes with a discussion of how the causal Bayesian approach deals with the criticisms of the standard Bayesian model and why it is superior to alternative approaches to reasoning about uncertainty popular in the AI community.

## 1. INTRODUCTION AND OVERVIEW

Medical diagnosis is one of the major foci of work on reasoning about uncertainty in artificial intelligence. It is the problem of how optimally to combine evidence from outwardly visible patient symptoms and signs to make the best inference about underlying or invisible disease causes, by using expert knowledge of the relative strengths of the links between causes and effects. The situation is complicated because each cause has multiple effects and several causes may produce the same effect. Also, certain effects are highly intercorrelated, so that treating them as independent may lead to diagnostic errors. Finally, the utility of treatment decisions varies among possible diseases, so that the most rational treatment decision does not necessarily treat the most probable disease. Effective methods of diagnosis should provide enough information to allow the best treatment choice for a given patient.

It may take a physician a lifetime to develop advanced diagnostic expertise in a particular medical specialty. Expert diagnosticians perform well because they have learned much about the placement and strengths of the links between causes and effects in the disease process (Clarke, 1982; Elstein, Shulman, & Sprafka, 1978). However, this does not imply that they have a superior calculus or set of production rules for reasoning about uncertainty. A large body of psychological research (recent reviews are Baron, 1985; Kahneman, Slovic, & Tversky, 1982; Nisbett & Ross, 1980) suggests that people reason about uncertainty using simplifying heuristics and strategies, which commonly lead to biased judgements. Physicians, as people, are subject to the same reasoning biases (Berwick, Fineberg, & Weinstein, 1981; Christensen-Szalanski & Bushyhead, 1981; Detmer, Fryback, & Gassner, 1978; Eddy, 1982).

If a physician's probabilistic knowledge is extracted and combined using a normative calculus, it can result in a diagnostic accuracy better than that of the physician (Clarke, 1984). (This "bootstrapping" phenomenon has a long tradition in psychology, see, for example, Slovic & Lichtenstein, 1971). Our goal as applied scientists, in designing an expert system, should be to use our best theories about normative reasoning to design an inference engine, which can be combined with a knowledge base obtained from an expert who has extended experience observing a domain. The purpose of this paper is:

1) to show why we think the causal Bayesian approach provides the most normative basis for

an inference engine for reasoning about uncertainty.

2) to give an example of an implementation.

## 2. STANDARD BAYESIAN ASSUMPTIONS AND CRITIQUES

Bayes theorem provides a good starting point for the design of an ideal inference engine because it has a firm foundation in the axioms of subjective probability (Savage, 1954). The standard form of Bayes theorem calculates the conditional probability of a cause (disease) given effects (symptoms) in terms of the probabilities of the diseases (priors) and the probabilities of the symptoms given the diseases (likelihoods). The formula, as applied to a single symptom, is as follows:

$$p(D_j/s) = \frac{p(s/D_j)\,p(D_j)}{\sum_{i=1}^{i=n} p(s/D_i)\,p(D_i)}$$

Here the D's refer to diseases, s is a particular symptom, and $D_j$ is the disease of interest.

The formula becomes more complicated for the conditional probability of a disease given more than one symptom. For example, the equation to find the posterior probability of disease $D_j$ when symptom $s_k$ is added to the k-1 known symptoms is:

$$p(D_j/s_1 \& s_2 \cdots \& s_k) = \frac{p(s_k/D_j \& s_1 \& s_2 \cdots \& s_{k-1})\,p(D_j/s_1 \& s_2 \cdots \& s_{k-1})}{\sum_{i=1}^{i=n} p(s_k/D_i \& s_1 \& s_2 \cdots \& s_{k-1})\,p(D_i)}$$

This means that to build a complete Bayesian system containing k symptoms and n diseases, it would be necessary to store likelihoods of each symptom given all combinations of from one to k-1 other symptoms with each of the n diseases. Therefore, traditional Bayesian diagnostic systems of sufficient scope to be useful in multiple disease situations are quite often difficult to implement, as has been repeatedly pointed out in reviews of the literature on computer-based diagnostic systems (Charniak & McDermott, 1985; Hayes-Roth, Waterman, & Lenat, 1983; Kleinmuntz, 1984; Schaffner, 1981).

### 2.1. ASSUMPTION OF SYMPTOM INDEPENDENCE

The practicality of Bayesian systems can be improved by assuming symptom independence, thereby reducing the size of the set of symptom likelihoods that need to be known beforehand. Symptom independence is expressed through the following two assumptions:

$$p(s_i/s_j) = p(s_i)$$
$$p(s_i/D \& s_j) = p(s_i/D)$$

The first says that the probability of observing symptom $s_i$ in the subpopulation of individuals with symptom $s_j$ is the same as the probability of observing symptom $s_i$ in the whole universe. The second says that the probability of observing symptom $s_i$ in the subpopulation of individuals with disease D and symptom $s_j$ is the same as the probability of observing symptom $s_i$ among all individuals with disease D. The upshot is these assumptions is that it is necessary to store likelihoods of only individual symptoms given diseases because the likelihood of two or more symptoms occurring together given a disease can be obtained by multiplying the individual likelihoods as follows:

$$p(D_j/s_1 \& s_2 \cdots \& s_k) = \frac{p(D_j) \prod_{l=1}^{l=k} p(s_l/D_j)}{\prod_{l=1}^{l=k} \sum_{i=1}^{i=n} p(s_l/D_i)\,p(D_i)}$$

The justification for these independence assumptions is questionable. Although the medical AI literature reports examples of independence-assuming Bayesian medical systems (reviewed in deDombal, 1979; Wardle & Wardle, 1978) having diagnostic accuracies of over 90%, it is apparent that symptoms are not usually independent of each other in the real world, because the probability of two symptoms of the same disease occurring together, given that one has the disease, is often much higher than the product of the individual probabilities of the symptoms. If someone has a headache, knowing that it was caused by a cold makes a simultaneous sore throat more likely. Monte Carlo studies by Russek, Kronmal, and Fisher (1983) imply that the assumption of symptom independence does not change the rank ordering of diseases, but it does effect the posterior probability distribution, so that the probabilities of diseases with p's over 0.5 are overestimated, and those with p's under 0.5 are underestimated. This would not be problematic for most computer medical diagnostic systems that just come up with lists of likely diseases, but has serious implications if one is trying to perform patient management by weighting utilities with accurate disease probabilities to find an optimum treatment.

### 2.2. ASSUMPTION OF SINGLE SYMPTOM CAUSES

In Bayes theorem causes of effects are held to be mutually exclusive. Each cause is assigned a prior probability or degree of belief, and when evidence is presented the degree of belief in a cause is incremented or decremented based on the ability of that cause to uniquely account for that evidence. When applied in the medical domain, the usual practice is to map causes onto single diseases. Single, rather than multiple (or single and multiple), diseases are chosen as causes because: people have a better understanding of the mechanisms, and hence probabilities, of symptoms being produced by single diseases; the necessity of keeping track of the probability of all symptom combinations given all disease combinations would make the combinatorial explosion even worse; and people have an understanding of the treatments that should be applied in the event of single, but not multiple, diseases. The problem with this assumption, like with the assumption of symptom independence, is that it violates real-world experience. Having two or more diseases co-occur in the same person is common, and treating one while ignoring the other may have serious medical consequences.

## 3. THE CAUSAL BAYESIAN MODEL

The causal Bayesian approach allows most of the computational simplicity obtained from assuming symptom independence, while providing a sensible way to model symptom interdependence. Symptoms of a disease that are intercorrelated are assumed to have an additional shared cause, besides the disease, which independent symptoms of the disease do not have. For example, anorexia (loss of appetite) and nausea, two symptoms of appendicitis, co-occur more frequently than would be expected by chance, because they both result from gastrointestinal disturbance. This additional cause, named a pathstate by Charniak (1983), has itself a certain likelihood of being caused by the disease. (Our notion is derived from Charniak, but similar ideas have also been proposed by Pearl, 1985.)

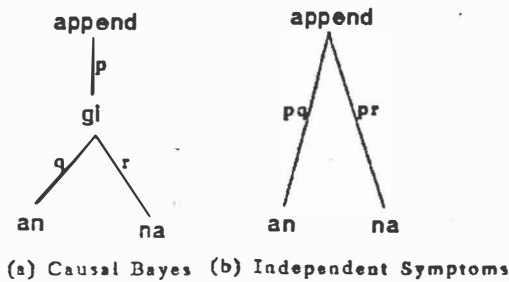

(a) Causal Bayes  (b) Independent Symptoms

Figure 1

The symptoms of gastrointestinal disturbance, anorexia and nausea, cannot appear until appendicitis has produced the disturbance, but after it is present, one or both symptoms may occur, relatively independently of the other. This is modeled by assuming that symptoms are independent with respect to proximal causes, which can be expressed computationally as follows:

p(anorexia & nausea/appendicitis) =
p(anorexia/gi disturbance) p(nausea/gi disturbance)
p(gi disturbance/appendicitis)

In other words, the likelihood of a set of symptoms given a disease is equal to the product of the likelihoods of the symptoms given the pathstate directly causing them times the likelihood of the pathstate given the disease. (The same principle applies, mutatis mutandis, if additional pathstates and symptoms are added.)

The contrast of the causal Bayesian approach with complete symptom independence is demonstrated by comparing Figures 1(a) and 1(b). In both figures, the likelihood of a gi disturbance given appendicitis is represented by p, the likelihood of anorexia given gi disturbance is represented by q, and the likelihood of nausea given gi disturbance is represented by r. According to the causal Bayesian approach the probability of anorexia and nausea both occurring given appendicitis would be pqr, while according to complete symptom independence it would be pqpr, making this unrealistically infrequent.

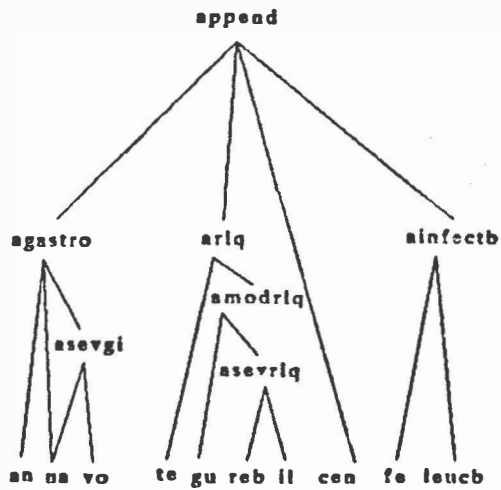

Figure 2

In our implementation a disease is represented by a hierarchical graph of a tree, expanding downwards, with the top node representing the disease, middle nodes representing pathstates, and bottom nodes representing symptoms. Figure 2 shows the representation we used for appendicitis. Links between nodes, which are only in a vertical direction, represent the probability of an child node being caused by its direct parent. All the child nodes which are direct descendants of the same parent are treated as probabilistically independent with respect to that parent. The likelihood of the vector of symptom nodes, $\vec{z_j}$, that are direct or indirect descendants of a node x having any particular configuration of

231

values, given x, is evaluated by the function "lhood" as follows:

If (x is a symptom node)

$$\text{Then lhood}(\vec{z}_j/x) := 1$$
$$\text{Else lhood}(\vec{z}_j/x) := \prod_{i=1}^{i=n} p(y_i/x) \text{ lhood}(\vec{z}_k/y_i)$$

Here, $y_i$ is a direct descendent of x, and $\vec{z}_k$ is the symptom vector of direct and indirect descendents of $y_i$. Individual symptoms can have values of present, absent, or unknown. Other than this new method for calculating likelihoods, the rest of Bayes theorem is used in the way described in the previous section.

## 4. THE KNOWLEDGE ENGINEERING

We started the project by locating an expert notable both for his medical knowledge and his sympathetic attitude towards attempts to quantify the medical decision-making process. We decided to work on the problem of the differential diagnosis of appendicitis because it was circumscribed, had hard data available, and the large differences in utilities for the possible treatment options would put a premium on the correct assessment of disease probabilities. The first author spent several weeks reading the literature of the domain, and after extended discussion we chose to limit the modeling process to the six diseases with prior probabilities higher than one percent that would be considered when diagnosing appendicitis. The final model also included 19 symptoms and 32 intermediate pathstates. (The hierarchical structures of the diseases were defined in PROLOG by the first author.)

We developed the pathstate structure for the various diseases by deciding which symptoms were causally related because they all were manifestations of a stress produced by the disease on a particular system, or organ, or at a particular location. In general, similar symptoms were interrelated in the same way for different diseases, simplifying the modeling process.

Although the diseases that are commonly confused with appendicitis may cause the same symptoms with the same probabilities, they can be diagnostically differentiated because they characteristically produce these symptoms at different times after the onset of the disease. Therefore, we needed to modify the pathstate models to allow diseases to evolve over time. We solved this problem by making the likelihood links between nodes conditional on the time since disease onset. This was done by having the expert graph the probability strength of each link as a function of time, from 0 to 132 hours. Even though drawing a line on a graph does not create a much larger time demand on the expert than making a point estimate, it provides much more information. In this way, we could generate likelihoods for symptoms being produced by the various diseases at whatever time since onset that the symptoms were reported. (Time since disease onset was defined as time since first observed symptom.)

Pathstates emerge from the expert's causal understanding of the disease process, but there is no independent evidence of their existence. As scientists, we include them as hypothetical constructs in our models because they improve the quality of our explanations and predictions. However we were posing our expert the difficult question of assigning probabilities to the likelihood of diseases causing theoretical entities or the theoretical entities causing symptoms. To reassure ourselves that he wasn't pulling these numbers from a hat, we devised a coherency check. We asked our expert to graph the direct likelihood of each symptom given appendicitis. These likelihoods should be both more reliable and better calibrated, because they are based on the empirical observation of symptoms and are consistent with the way our expert is used to thinking about diseases. We could then compare these likelihoods with the likelihood of a symptom given a disease obtained from the model and correct the discrepancies. (The model defines the likelihood of a symptom given a disease as equal to the products of the likehoods linking all the nodes on a direct path from the symptom to the disease.) There was substantial overlap, although the model estimates tended to be lower than the direct ones (perhaps because the expert underestimated the effect of multiplying several likelihoods less than one when providing the original pathstate probabilities).

The knowledge engineering also involved obtaining estimates for disease priors and treatment utilities from our expert. We found the priors to be conditional upon age and sex, so we repeated the technique of having the expert draw graphs to show the change in disease probability with age. Some of the prior probabilities of the gynecological diseases also varied according to time of the month, so we also made a graph of these changes and used them to weight the age-based priors. (Afterwards, the prior probabilities across diseases were normalized.) Entering treatment utilities involved the related problems of finding one standard to measure utility and defining the treatment options. The eventual options used were symptomatic treatment until recovery or an intraperitoneal operation (symptomatic treatment implies that an operation may be performed if the patient's condition declines). Utility was estimated in terms of morbidity, specifically, number of days in the hospital expected for a particular disease-treatment combination (mortality was not used as a measure of utility, because it is typically below one percent for all cases).

## 5. IMPLEMENTING THE MODEL

The rest of this section describes several complications that emerged when we tried to implement the model and the solutions we arrived at to resolve them.

## 5.1. MODELING DIFFERENT LEVELS OF SEVERITY

One problem we encountered was that symptoms could be related without being simply independent with respect to one another, because they reflected different degrees of severity of the same stress. We solved this problem by adding an additional pathstate below the main one to represent each increasing level of severity that we wanted to model. Each pathstate had some probability of causing a direct descendant pathstate reflecting the next higher degree of severity. Pathstates at both levels could also directly cause symptoms, though not usually the same ones. For example, in Figure 2, the pathstate *arlq* reflects the presence of any right lower quadrant peritoneal signs including the most minor, tenderness. If the peritoneal signs reach moderate intensity, reflected by *amodrlq* being caused, then there should also be a high probability of observing guarding. Finally, if the right lower quadrant peritoneal signs reached their most severe, shown by *asevrlq*, then rebound tenderness and ileus should appear as well. The advantage of this kind of structure is that it is impossible for the severe signs of a stress to occur without the mild ones, because the mild ones are caused by a pathstate directly caused by the disease. (We would also expect the low severity pathstates and their symptoms to be present earlier than the high severity pathstates and theirs, because increasing severity evolves over time. This is relatively easy for us to model, because, as stated in the previous section, the likelihood links between pathstate nodes are conditioned on time.)

## 5.2. COMPENSATING FOR SINGLE CAUSES OF SYMPTOMS

Another serious problem to emerge was that not all symptoms we were looking at were caused by all the diseases in the hypothesis pool. Theoretically, the observation of a symptom uniquely caused by one of the diseases should categorically rule out the others. No doctor would perform a diagnosis on this basis. We might try to deal with this problem by relaxing the condition that diseases be mutually exclusive and by adding other diseases thought to cause the symptoms, e.g., colds. We resisted this approach. It seemed reasonable to consider our six diseases to be mutually exclusive for the practical purpose of differentially diagnosing *appendicitis*. Although any one of the symptoms we were considering might be caused by many different diseases, the co-occurrence of several of these symptoms in the patient would strongly suggest that the relevant disease was from the pool. And since the pool was so small, we felt that the chance of two diseases from within it occurring simultaneously to cause acute symptoms to be vanishingly small.

The solution was to allow the possibility that symptoms could be independently caused by individual diseases from outside the pool and were co-occurring with the diseases from within it. Accordingly, we had our expert graph a "base-rate" external-cause probability for each symptom by sex and age. The numerator of Bayes theorem would now become:

$$p(D_i) \text{ lhood}(\vec{z}_j^*/D_i) \prod_{l=1}^{l=n} p(z_l)$$

i.e., the prior probability of the disease times the likelihood of the symptoms caused by the disease having a particular set of values, given the disease, times the independent probability of each symptom not caused by the disease having its particular value. $D_i$ represents any disease from the hypothesis pool, and $z_l$ stands for any symptom included in the model not directly or indirectly caused by $D_i$. The denominator becomes the sum of these expressions across diseases.

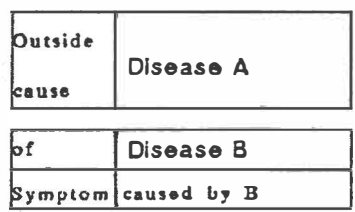

Figure 3

There are two ways that "base-rate" symptoms could co-occur with the diseases of the hypothesis pool and their symptoms, as shown in Figure 3. The entire upper rectangle in Figure 3 represents the probability of Disease A, and the entire lower rectangle represents the probability of Disease B. Disease B causes a symptom with a certain likelihood, represented by the lower half of Disease B, while Disease A does not cause it. There is also a certain probability of this symptom co-occurring independently with each disease from outside causes, represented by the small vertical rectangles on the left sides of A and B. The probability of this symptom given Disease A is simply equal to the symptom's overall "base-rate" probability, because Disease A and the symptom are independent. The probability of the symptom given Disease B involves two contributions, one from inside the disease and one from outside it. Because they are independent contributions, the total probability of the symptom given Disease B is equal to the probability of the symptom being caused by the disease, plus the probability of its being caused externally, minus the overlap.

The same principle can be extended for each symptom that either is caused or not caused by a particular disease. However, for a symptom to be caused by a disease, there is a long causal chain to follow, any of whose links could break. In other words, a symptom could be absent just as easily from a pathstate not being caused by a disease as from a symptom not being caused by a pathstate. Conversely, any time there is a

233

break in the causal chain, there is some probability that a given symptom could be independently caused. Therefore the revised "lhood" function should include a probability for an independent symptom to be observed given all possible outcomes for the causal chain, either being intact or breaking at any given link. It is as follows:

If (x is a symptom node)

   Then $\text{lhood}(\vec{z}_j/x) := 1$

   Else $\text{lhood}(\vec{z}_j/x) :=$

$$\prod_{i=1}^{i=n} [(p(y_i/x)\, \text{lhood}(\vec{z}_k/y_i)) + ((1 - p(y_i/x)) \prod_{k=1}^{k=m} p(z_k))]$$

Here $p(z_k)$ is the externally caused probability of a symptom descendant of node $y_i$ having a value of present, absent, or unknown. The "base-rate" probability assigned for a symptom $z_k$ being absent is one minus the probability $p(z_k)$ of that symptom being present, and the value assigned for unknown status is one (so that the product in the numerator of Bayes theorem depends only on known symptoms). The term on the left side of the definition of "lhood" for disease and pathstate nodes remains unchanged from the previous definition. The term on the right side represents the probability of the set of descendant symptoms being observed in the absence of node $y_i$ being caused.

### 5.3. COMBINING INFORMATION FROM TWO MEASUREMENTS

### 6. TESTING THE MODEL

We tested the ability of the model to distinguish the symptoms of 100 recorded cases of appendicitis from those of 100 recorded cases of nonspecific abdominal pain. The cases were chosen to present a wide range of diagnostic difficulty, with some causing high probabilities of error for human diagnosticians and previous (Bayesian) diagnostic systems. For the test, we constructed a comparison independent Bayesian model which had the same likelihoods for individual symptoms being caused by a disease, but in which the probabilities of combinations of symptoms given a disease were obtained by multiplying their separate likelihoods, as in Figure 1(b). The probabilities of individual symptoms for the independent model were found by multiplying all the likehoods linking all the nodes on a direct path from the symptom to the disease in the pathstate model. This would provide a fair comparison; any difference in the performance of the models could not be attributed to the accuracy of the likelihoods of individual symptoms and must therefore depend on how the data about symptoms were combined.

To accurately describe the test sample, we set the prior probabilities of appendicitis and non-specific abdominal pain to 0.5 in both models. We entered the symptoms of the cases into both models and obtained posterior probabilities of appendicitis for each case. An appropriate measure of performance is calibration, also known as "reliability in the small" (Yeats, 1982). A person or decision-system is well calibrated if the probability she or it assigns to a given outcome occurs that with that frequency; for example, if 9 out of 10 abdominal cases that were assigned a 90% probability of of being appendicitis actually were appendicitis. (We are more interested in calibration than in maximizing discrimination between appendicitis and non-appendicitis because good calibration allows maximum accuracy in placing diseases above or below the probability threshold for switching treatment options based on utility considerations. This threshold is rarely 0.5).

The result, obtained using the jackknife statistical technique (Mosteller & Tukey, 1977), was that the calibration of the causal Bayesian model (.0735) was superior to the independent model (.0785). The probability of this difference being due to chance was <.001. An alternative comparison using a measure of the area of error between a quadratic regression function fitted on the probabilities of the cases assigned by a particular model and the perfect calibration function also showed the causal model to be significantly superior. The results suggest that the causal Bayesian approach can provide a viable solution to the interdependence problem.

These tests are described in more detail in Schwartz, Clarke, Baron, & deDombal (1986). A test of a conceptually similar causal Bayesian model which used objective probabilities to define the links between pathstates, but which did not allow for outside causes of symptoms, also showed a considerable improvement in calibration over an independent comparison model (Clarke, Schwartz, Baron, & deDombal, 1986).

### 7. WORK IN PROGRESS

Another problem that we have been considering is how to take advantage of symptom information from two different times. For example, a doctor learns about a first set of symptoms, noticed by the patient when he contracts a disease, and a second set of symptoms when the patient is examined. Between the two measurements, symptom values may change in either direction. Given that one has a model of how various diseases evolve over time, these changing symptoms should be very diagnostic in distinguishing among diseases.

Suppose the likelihood of a symptom given a disease increases over time. Four symptom patterns may be observed: yes-yes, yes-no, no-yes, or no-no. The two temporal measurements are not independent, because they reflect the operation of a single disease process. Probabilities may be assigned to these patterns by assuming a relationship of implication between likelihoods. If a symptom is caused by a disease at a time when it is less likely to do so, then we must also expect it to occur when it is more likely. However, a symptom

234

occurring when it is more likely does not necessarily imply that it will occur when it is less likely. Therefore, the probability of the yes-yes pattern may be defined as probability of the symptom at the less likely time. The yes-no pattern, with a symptom being observed at the less likely, but not more likely, time, is impossible and could only be produced by some outside cause. The no-yes pattern, with the symptom being observed only at the more likely time, has a probability equal to the difference between the likelihoods of the symptom at the more and less likely times. The probability of the no-no pattern is equal to one minus the likelihood of the symptom being caused at the more likely time.

This logic can easily be extended to describe the temporal changes in the likelihood of any child node given its parent. A parallel representation describes how the four patterns can be accounted for by independently caused symptoms. Work is proceeding to complete the implementation of these additions and test them.

At the same time, the first author is using the causal pathstate model of appendicitis described in the previous sections of this paper to compare the diagnostic reasoning of expert and novice surgeons, as part of his doctoral dissertation in Psychology at the University of Pennsylvania.

## 8. CONCLUSIONS

There are several conclusions we would like to draw about the work we have described. We have tried to show that a Bayesian diagnostic system using subjective causal links between intermediate states is feasible. Above that, the calibration results show that causal modeling is consequential, that is, it visibly improves performance over a Bayesian system based on the conventional assumption of symptom independence. Once again, the lesson of Artificial Intelligence research is that a simple theoretical analysis of a problem does not reveal all the difficulties that arise during implementation. We have found additional difficulties, not foreseen by Charniak (1983), specifically:

1) that symptoms take on different meanings because diseases evolve over time.
2) that symptoms can be indirectly related, because they reveal different levels of severity of a pathological condition.
3) that not all the diseases in an hypothesis pool cause the same symptoms, so that it is necessary to allow symptoms to have external causes.

Fortunately, these problems are soluble.

The main advantage of this approach that we would like to emphasize is that unlike the current "hot" approaches for dealing with uncertainty in AI, MYCIN's certainty factors and Dempster-Shafer (Buchanan & Shortliffe, 1984; Shafer, 1976; Shortliffe, 1976), the output of this system is real probabilities that can be used to weigh the utilities of teatment options for patient management. Other advanteges are that it solves the interdependence problem in a palatable way, is computationally simple, and does not place an excessive demand for probability estimates on the expert. Given the theoretical justification for the Bayesian approach, there now seems to be no further excuse for neglecting it in the design of expert systems.

Acknowledgements: Thanks to Ruzena Bajcsy for her unending generosity and intellectual and spiritual support without which this work would not have been possible.